%% file: main.tex
\pdfoutput=1

\documentclass[11pt]{article}

\usepackage[]{acl}

\usepackage{times}
\usepackage{latexsym}

\usepackage[T1]{fontenc}

\usepackage[utf8]{inputenc}

\usepackage{graphicx}
\usepackage{amsmath}
\usepackage{amssymb}
\usepackage{amsbsy}
\usepackage{textcomp}
\usepackage{bbm}
\usepackage{bm}
\usepackage{multirow}
\usepackage{tabularx}
\usepackage{ragged2e}
\usepackage{booktabs}
\usepackage{subfig}
\usepackage{tablefootnote}
\usepackage{footnotehyper}
\usepackage{xspace}
\usepackage{enumitem}

\DeclareMathOperator*{\argmax}{arg\,max}
\DeclareMathOperator*{\argmin}{arg\,min}

\usepackage{microtype}
\usepackage{stmaryrd}

\newcommand{\kl}[1]{\textsc{kl}\left(#1\right)}

\newcommand{\chaosNLIs}{ChaosNLI-S\xspace}
\newcommand{\chaosNLIm}{ChaosNLI-M\xspace}
\newcommand{\chaosNLIa}{ChaosNLI-$\alpha$\xspace}
\newcommand{\abdnli}{$\alpha$NLI\xspace}
\newcommand{\pavnli}{PK2019\xspace}

\title{Distributed NLI: Learning to Predict Human Opinion Distributions for Language Reasoning}

\author{Xiang Zhou\thanks{\,\, Equal contribution.}\ \ \ \ \ \ Yixin Nie\footnotemark[1]\ \ \ \ \ \ Mohit Bansal \\

  Department of Computer Science \\
University of North Carolina at Chapel Hill \\
  \texttt{\{xzh, yixin1, mbansal\}@cs.unc.edu} \\
}

\date{}

\begin{document}
\maketitle

\input{0abstract}

\input{1introduction}

\section{Distributed NLI}
\label{sec:prelude}
\subsection{Natural Language Inference}
NLI was first introduced and mostly formulated as a 3-way classification problem. The input is a premise paired with a hypothesis. The output $y$ is a discrete and mutually exclusive label that can be entailment, neutral, or contradiction, indicating the truthfulness of the hypothesis given the premise.
Some works advocated a shift for NLI from the 3-way discrete labeling schema to a graded schema due to the probabilistic nature of entailment inference~\cite{zhang2017ordinal, chen-etal-2020-uncertain}. Following such schema, models were instead required to produce a continuous score representing how likely the premise is true given the hypothesis. No matter whether the label is discrete or graded, the conventional goal of NLI in most recent literature is to develop models to make the inferences that an \textit{individual} would naturally make with an implicit assumption that there is only one true label.

\subsection{Task Definition}

We introduce distributed NLI by extending the conventional NLI label to be a distribution representing collective human opinions on the example.
Specifically, the goal of distributed NLI is to \textbf{\textit{develop NLI models that can predict a categorical distribution similar to the real human opinion distribution obtained from a large population.}} 
In the following subsection, we explain the motivation and importance of distributed NLI. 

\begin{table*}[ht]
\centering
\tiny
\scalebox{0.97}{
\begin{tabular}{p{19em}p{17em}p{8em}p{25em}}
\toprule
{\bf Premise} &  {\bf Hypothesis} &
{\bf Labels} &  {\bf Hypothetical Reason for the Disagreement} \\
\midrule
To savor the full effect of the architect's skill, enter the courtyard through the gate which opens onto the Hippodrome. & The gate to the Hippodrome is an example of the architect's skill. & E\textsuperscript{(76)} N\textsuperscript{(22)} C\textsuperscript{(2)} & Annotators might have different judgements on what is demonstrating the architect's skill. The gate is highly possible for some annotators but it is not certain for others.\\
\midrule
Look, there's a legend here. & See, there is a well known hero here. & E\textsuperscript{(57)} N\textsuperscript{(42)} C\textsuperscript{(1)} & Whether ``a legend'' refers to a ``well known hero'' is debatable and subjective. \\
\midrule
While it's probably true that democracies are unlikely to go to war unless they're attacked, sometimes they are the first to take the offensive. & Democracies probably won't go to war unless someone attacks them on their soil & E\textsuperscript{(66)} N\textsuperscript{(31)} C\textsuperscript{(3)} & The words like ``probably" and ``sometimes" make it hard to determine whether the ``democracies" will be the first to attack or not. \\
    
\bottomrule
\end{tabular}
}
\vspace{-3pt}
\caption{We show 3 examples from \chaosNLIm with their distribution labels and our hypothesis regarding how the disagreement arises.}
\vspace{-6pt}
\label{tab:dist_example}
\end{table*}

\subsection{Motivation and Positioning}
\label{sec:motivation}
Advocated by~\citet{you2006local}, annotation tasks of NLI should be ‘‘natural’’ for untrained annotators, and that the role of NLP should be to model the inferences that humans make in practical settings without imposing a prescriptivist definition of what types of inferences are licensed.\footnote{There has
been a gravitation towards the preference of natural inference over rigorous annotation guidelines based on a prescriptive definition of entailment relation in logic. We refer readers to~\cite{pavlick2019inherent} for a more comprehensive discussion on the topic.}
Maintaining the ``naturalness'' of inference instead of referring to a strict definition of logic entailment facilitates the practical usage of NLI, however, it unfortunately brings a degree of uncertainty to the inference among different individuals.
Recent findings reveal that inherent disagreements exist in a noticeable amount of examples in oft-used NLI datasets~\cite{pavlick2019inherent, nie2020can}.
Hence, the conventional goal of NLI (i.e., to model the natural thinking process of a single human) may have a risk of ill-definition because a consensus on the label cannot be reached for some cases.\footnote{In such cases, we cannot coerce a most legitimate label by giving a prescriptivist definition of the inference since it will contradict the ``naturalness'' of the task described above.} Examples are shown in Table~\ref{tab:dist_example}. 
Moreover, with such label agreements, traditional evaluation methods using a single label may also become unreliable~ \cite{gordon2021disagreement}.
Our proposed distributed NLI resolves such a risk without compromising the naturalness of the inference.

An alternative approach toward the inherent disagreements is to narrow the task to model only the majority label. This is the default setup for most prior studies where multiple labels were collected for the examples in the development and test sets and the majority label will be chosen as the gold label upon which the accuracy will be calculated. We argue that such a practice is insufficient.
With the advancement in general language modeling for NLU, we could envision NLI models having a potential influence on AI-aided critical decision making.
Such decisions may be involved when assisting a jury's verdict of a lawsuit given the vocal and textual reports about the case~\cite{surden2019artificial,armour2020ai}, providing automated opinions for company recruiting or university admissions based on personal information~\cite{Ochmann2020AIRE,newman_frazier_miller_2020}, or even helping governments make decisions~\cite{eggers_schatsky_viechnicki_2017} (see the Appendix~\ref{app:application} for potential NLI inputs for these applications). 
Hence, it would be important for the system used in such a decision-making process to be aware of different opinions and to pass the distribution of the collective opinions to either the actual decision maker or any downstream models. 

\paragraph{Merits of Distribution Labels.}
\label{app:merits}
The Distributed NLI and the traditional format of NLI seem to be two similar tasks with a major difference as using the distribution labels instead of the one-hot labels. However, we argue that these distribution labels can capture more fine-grained and subtle semantics that may have a great impact on downstream applications, which is ignored in the traditional one-hot labeling schema.  
Firstly, \textit{distribution modeling captures more semantic subtleties.}
Three examples are shown in Table~\ref{tab:dist_example}. In order to predict the corresponding label correctly, the model needs to understand all these challenging language properties, including ambiguous relationships between phrases (e.g. ``legend'' vs. ``well known hero'' in the second example), sentences with subjective understandings (the first example), sentences with more complicated relationships hard to attribute any of the three classes conclusively (the third example). These challenges are not visible in the traditional one-hot label schema, but become essential to solving the distributed NLI task. 
Additionally, \textit{capturing these semantic ambiguities can also lead to great impact in downstream tasks and real-life applications}. NLI models are widely used in various downstream tasks either to conduct a sub-step and to provide rewards~\cite{pasunuru-bansal-2017-multi,falke-etal-2019-ranking}, where the data distribution can be diverse and noisy (e.g. model-generated sentences are usually imperfect), which leads to more complex and ambiguous labels. Models capturing better label distribution can be more useful in these downstream applications, as well as in the potential decision-making applications.

\paragraph{Remark on Labeling Schema.} For the study of distributed NLI in this work, we maintain the discrete labeling schema rather than the graded labeling schema because this is the default format by which most of the natural data is recorded. The discrete label is also more straightforward for annotation, since annotators are accustomed to providing their discrete judgement (yes or no, true or false) in daily life, but usually not a real value indicating how confident (or strong) their feelings are.  Note that despite the schema choice in this work, the concept of distributed NLI can be easily generalized to graded-label settings where the target is to fit a distribution of the continuous grade score. Finally, there can be a connection between the distributed NLI categorical distribution and the graded score annotated by an individual human. Despite their different meanings, the judgement of individual humans can sometimes be influenced by their belief of other people's thoughts~\cite{kovacs2010social}.

\paragraph{Remark on Future Directions.} Additionally, an ideal model should also be able to capture the detailed thought process behind the prediction of each label and provide corresponding explanations. Such interpretability will make the model more reliable in critical applications, but is generally beyond the capability of current models and hard to evaluate under current datasets. While related information can be extracted from current models by using post-hoc interpretability tools (e.g. LIME~\cite{ribeiro2016should}), we encourage future works to build more interpretable models and collect datasets suitable for more fine-grained evaluations.

\paragraph{Remark on Annotation Quality.}
Evaluation of distributed NLI compares model prediction to the opinion distribution estimated by multiple annotations. We noticed that examples with a high-level of disagreement usually require more mental effort to annotate. 
While previous work~\cite{pavlick2019inherent,nie2020can} have conducted analyses showing these collected label distribution contain genuine intrinsic disagreement, we also notice unreasonable labels that may just come from annotation noises. So far, it is still unclear whether the collected distribution labels are \textit{high-quality} and \textit{clean} enough to serve as evaluation datasets.
Therefore, to ensure that the evaluation is valid, it is crucial to maintain the quality of annotations such that the label distribution will indeed represent opinion diversity rather than annotation errors. As an example, we use ChaosNLI~\cite{nie2020can} in our experiments, which is collected with careful quality control.\footnote{Note that despite the three-way discrete label schema choice in this work, the concept of distributed NLI can be easily generalized to other datasets, including graded-label settings where the target is to fit a distribution of the continuous grade score. More discussion is in the Appendix~\ref{app:motivation}} Furthermore, we conducted a manual quality check on 100 examples from the \chaosNLIm (the $D^{dev}_s$ subset later to be introduced in Sec. \ref{sec:task}). Each example has 100 three-way annotated NLI labels, and we examine whether any of the 100 annotations for each example will be an absolute error in almost all scenarios. In total, only 4 (out of 100) examples contain more than 10 error annotations and no example contain more than 16 error annotations.
Quantitatively, we have also verified that these errors do not substantially impact the findings and comparisons in this paper. More detailed results and examples on annotation quality analysis are in the Appendix~\ref{app:quality}.

\section{Dataset and Experiment Design}
\label{sec:task}

In this section,  we describe a typical design of dataset and experiment of the distributed NLI task, and is used in later experiments in this work.
For a typical NLI task, the dataset is split into train, development, and test set where each example is associated with one ground truth label. The model will be trained using examples in the training set. Accuracy on the development set is used for model selection, and accuracy on the test set will be reported as the final metric. For distributed NLI, in order to develop models that can predict the label distribution,
we assume that each example in the test set will also have a sufficient amount of human labels to approximate the real human label distribution to evaluate the model's prediction.

Let us define the $D^{train}$, $D^{dev}$, $D^{dev}_s$, and $D_s^{test}$ to be the different splits of the dataset. The subscript $s$ in $D^{dev}_s$ and $D_s^{test}$ indicates that the examples in these two splits have soft labels representing the human label distribution, while there is no such label in $D^{train}$ and $D^{dev}$. $D_s^{dev}$ is a very small set of examples with soft-labels besides the test set. 
This gives a good simulation for real production because in practice, $D^{dev}_s$ will be extremely scarce. 
The goal of the distributed NLI is to develop models that can predict human label distributions and minimize the average divergence between predicted label distributions and approximated human label distributions on the test set using examples in $D^{train}$, $D^{dev}$, and $D^{dev}_s$. 
\footnote{Although $D^{dev}_s$ is also called development set, there is no strict relation between examples in $D^{dev}_s$ and $D^{dev}$. They could share some examples or can be mutually exclusive.}
Even though obtaining these soft label distributions is expensive, our design can generalize to the situation where we can also have enough training data with soft-label by simply making a new split $D^{train}_s$ on which the model can be trained.

\input{3estimate}
\input{4settings}

\input{5results}
\input{6analysis}

\input{7related}
\input{8conclusion}

\section{Ethical Considerations}
The main target of this paper is to propose a new extension of the NLI task that focuses on predicting the whole distribution instead of one single label. Our new formalization can potentially make the related application of NLI more reliable in the practice, as the models trained on our proposed task will focus more on minority opinions which may be ignored in the traditional formalization. Nonetheless, we are strongly against the use of current NLI models in any critical applications (e.g. admission, jury, etc.). While NLP models can use to help human judgment (and the results should be verified by a human), their robustness and fairness are still an unsolved issue, and cannot replace the work of human experts. 

\section*{Acknowledgments}
We thank Izzeddin Gur and the reviewers for their helpful comments.
This work was supported by ONR Grant N00014-18-1-2871, DARPA YFA17-D17AP00022,
NSF-CAREER Award 1846185, and DARPA MCS Grant N66001-19-2-4031. The views are those of the authors and not of the funding agency.

\bibliographystyle{acl_natbib}
\bibliography{main}

\input{appendix}

\end{document}

%% file: 0abstract.tex
\begin{abstract}

We introduce distributed NLI, a new NLU task with a goal to predict the \textit{distribution} of human judgements for natural language inference. 
We show that by applying additional distribution estimation methods, namely, Monte Carlo (MC) Dropout, Deep Ensemble, Re-Calibration, and Distribution Distillation, models can capture human judgement distribution more effectively than the softmax baseline.
We show that MC Dropout is able to achieve decent performance without any distribution annotations while Re-Calibration can give further improvements with extra distribution annotations, suggesting the value of multiple annotations for one example in modeling the distribution of human judgements. 
Despite these improvements, the best results are still far below the estimated human upper-bound, indicating that predicting the distribution of human judgements is still an open, challenging problem with a large room for improvements. We showcase the common errors for MC Dropout and Re-Calibration. Finally, we give guidelines on the usage of these methods with different levels of data availability and encourage future work on modeling the human opinion distribution for language reasoning.\footnote{Our code and data are publicly available at \url{https://github.com/easonnie/ChaosNLI}.} 
\end{abstract}

%% file: 1introduction.tex
\section{Introduction}
Natural Language Understanding (NLU) and Reasoning play a fundamental role in Natural Language Processing (NLP) research. It has almost become the de facto rule that newly proposed generic language models will be tested on NLU tasks and progress obtained on general NLU often bring potential improvement on other aspects of NLP research~\cite{wang2018glue}. The well-known NLU tasks include Sentiment Analysis~\cite{socher2013recursive}, Natural Language Inference (NLI)~\cite{bowman2015large, nie-etal-2020-adversarial}, Commonsense Reasoning~\cite{talmor-etal-2019-commonsenseqa}, etc., covering a representative set of problems for NLP.

One common practice shared by most of the language understanding and reasoning tasks is that they are formalized as a classification problem, where the model is required to predict a single most preferable label from a predefined candidate set, and the goal is to reverse-engineer how a reasonable human chooses the best one. This simplification not only helps standardize the evaluation, i.e., accuracy could become the canonical measure, but also help make the annotation task more straightforward during crowdsourcing data collection.

However, recent findings suggest that inherent disagreements exist in both the Natural Language Inference (NLI) and Commonsense Reasoning datasets~\cite{pavlick2019inherent, chen-etal-2020-uncertain, nie2020can} and advocate that NLU evaluation should explicitly incentivize models to predict distributions of human judgments. Similarly, \citet{gantt-etal-2020-natural} suggest that NLI should account for annotator random effects. This is intuitive since there might be different subjective views of the world and people might think differently given the same reasoning task especially those involving pragmatic reasoning~\cite{potts2016embedded}. Modeling the distribution of human opinions provides a higher level ``meta-view'' of the collective human intelligence which would be valuable for all aspects of NLP applications. 

In this work, as a case study for learning the distribution of human judgements on NLU, we extend the NLI task to distributed NLI -- a new task in which models are required to predict the distribution of human judgements for natural language inference. 
We introduce the new task based on the data from prior works~\cite{pavlick2019inherent, nie2020can} with new experimental guidelines designed for the distribution annotations. Standard NLP models are trained towards predicting single labels, while in theory models trained on single labels should still be able to capture the whole label distribution (see Appendix~\ref{app:deterministic} for a more detailed discussion), their predicted distribution may not be reliable~\cite{guo2017calibration}.
To achieve better distribution estimation and to maintain the merits of SOTA models, we consider four distribution estimation methods that do not need major architecture changes, namely, MC Dropout~\cite{gal2016uncertainty}, Deep Ensemble~\cite{lakshminarayanan2017simple}, Re-Calibration~\cite{guo2017calibration}, and Distribution Distillation for distributed NLI. These methods have achieved empirical success in estimating the aleatoric uncertainty~\cite{gal2016uncertainty}, calibrating the neural network prediction confidence~\cite{guo2017calibration}, and neural network knowledge distillation~\cite{hinton2015distilling}, respectively. We show that all four methods can substantially outperform the baseline and that Re-Calibration and Distribution Distillation can provide further improvement by making use of additional distribution annotations.
Specifically, our primary contributions are:
\begin{itemize}[leftmargin=*]
\vspace{-6pt}
\setlength\itemsep{-0.1em}
\item We introduce and define the distributed NLI task with the goal to model the distribution of human opinions on NLI. We also elaborate the motivation, feasibility (Sec.~\ref{sec:prelude}) and the experiential design (Sec.~\ref{sec:task}) for the task, serving as common ground for future research on the topic.
\item We test 4 methods (MC Dropout, Deep Ensemble, Re-Calibration, Distribution Distillation) for predicting the distributions over human judgments on NLI according to our experimental design, and find: (1) all methods bring substantial improvements over baseline; (2) Re-Calibration, MC Dropout, and Distribution Distillation are able to further improve the performance by using additional distribution annotations
(3) the best results are still far below the estimated human performance.
(4) MC Dropout and Re-Calibration can achieve decent generalization performance on out-of-domain distributed NLI test set without in-domain training data (Sec.~\ref{sec:results}).
\item Despite the improvement, we showcase common errors of MC Dropout and Re-Calibration and give guidelines on selecting methods and setting hyperparameters in different scenarios and argue for future work on modeling human opinions on language reasoning (Sec.~\ref{sec:analysis}).
\end{itemize}

%% file: 3estimate.tex
\section{Distribution Estimation}
The output of a typical NLI classifier is a vector $\mathbf{z} \in \mathbf{R}^3$ whose elements $z_i$ represent the unnormalized scores (or logits) for each of the three labels~\cite{parikh2016decomposable, nie2017shortcut}. In modern NLI models, the classifier is usually a deep neural network and the final output is:
\vspace{-6pt}
\begin{equation*}
    \hat{\mathbf{y}} = \text{Softmax}({\mathbf{z}}),  \quad \hat{c} = \argmax(\mathbf{z})
\vspace{-6pt}
\end{equation*}
where $\hat{\mathbf{y}}$ is the normalized label distribution whose element $\hat{y}_i = {e^{z_i}} / {\sum{e^{z_i}}}$, and $\hat{c}$ is the predicted label. Prior works~\cite{pavlick2019inherent, nie2020can} revealed that the distribution $\hat{\mathbf{y}}$ produced by the softmax layer gives a poor estimation on the real human label distribution. 

In this work, we experiment on using distribution estimation methods for predicting human opinion distribution on NLI, and we show that they can achieve better performance than the softmax output. These methods have been used in uncertainty estimation and confidence calibration with some empirical success. 
Although the problem of uncertainty estimation is different from opinion distribution estimation, the essence of the two are the same -- the estimation of a distribution.\footnote{Conceptually, capturing the distribution label in NLU tasks is similar to modeling the aleatoric uncertainty~\cite{kendall2017uncertainties}. And the uncertainty estimation of the opinion distribution can be itself a new task out of this paper's scope.}

\subsection{Bayesian Inference}
The Bayesian view of neural networks~\cite{mackay1992practical, neal1995bayesian} offers a mathematically
grounded framework to produce a distribution for the end task.
From a Bayesian perspective, we have a prior over possible models $p(\theta)$, a likelihood of the data $p(D | \theta)$, and we can use the expected posterior prediction as the final prediction distribution:
\vspace{-5pt}
\begin{equation*}
    p(\hat{y}|x) = \int_\theta p(\hat{y}|x, \theta) p(D|\theta) p(\theta) d\theta
\vspace{-1pt}
\end{equation*}
In practice, the integral over $\theta$ is intractable. We can approximate it by Monte Carlo sampling~\cite{metropolis1949monte} $\theta$ from an approximated posterior $p(\theta | D) \propto p(D|\theta)p(\theta) $ and then averaging their outputs.
\vspace{-5pt}
\begin{equation*}
    p(\hat{y}|x) = \mathbb{E}_{p(\theta | D)} \llbracket p(\hat{y}|x, \theta) \rrbracket \approx \frac{1}{k} \sum_i^k p(\hat{y}|x, \theta_i)
\vspace{-1pt}
\end{equation*}
where $\theta_i$ is one of the $k$ models sampled from the posterior $p(\theta | D)$. The calculation of the real posterior $p(\theta | D)$ is also intractable and there are multiple ways to approximate the model parameters sampled from the posterior. In this work, we consider two simple and empirically effective methods.

\paragraph{Deep Ensemble.}
The ensemble of neural networks~\cite{lakshminarayanan2017simple} has an intuitive Bayesian interpretation: network initialization is a sample from the prior $p(\theta)$ and network training is maximizing the data likelihood $p(D | \theta)$. Hence, sampling $k$ models from posterior $p(\theta | D)$ can be approximated by training $k$ models with different initialization and example ordering.

\paragraph{Monte Carlo Dropout.}
Sampling models by ensemble is computationally expensive because in total, $k$ models need to be trained, and even training one single model is already expensive for some tasks. Alternatively, \citet{pmlr-v48-gal16} proposed an efficient method that directly draws the samples by making $k$ stochastic forward passes with dropout in one single fully trained neural network. Loosely speaking, this is similar to obtaining samples by adding noise to a fully trained network~\cite{srivastava2014dropout}: $p(\hat{y} | x, \theta_i) = p(\hat{y} | x, \theta + \sigma_i)$. 
The method was shown to have good performances on neural network uncertainty estimation, and we refer the readers to the original paper for a detailed theoretic description.

\paragraph{Remark.} 
The Bayesian approach for the estimation of the NLI human label distribution has an appealing analogy to collective thinking. Sampling $\theta_i$ from parameter space can be seen as sampling an individual person from a large population with potentially diverse opinions. The stochasticity in personal experience is analogous to the randomness of network initialization and training dynamics.

\subsection{Re-Calibration}
The Bayesian method has a nice theoretical ground and does not require additional soft-labeled data $D^{dev}_s$. However, the empirical performance of Bayesian methods can be suboptimal due to overly idealized approximation.
Therefore, we also consider the method of calibration for distribution estimation which makes empirical post-editing on the output of the network by explicitly taking advantage of additional soft-labeled data $D^{dev}_s$. 
The core of calibration is to seek a proper scaling of $\mathbf{z}$ such that the calibrated output $\hat{y}$ can better present the objective distribution.
In our work, the calibrated predicted distribution is:
\vspace{-5pt}
\begin{equation*}
    \hat{y}_i=\frac{z_i/T}{\sum_i z_i/T}
\vspace{-3pt}
\end{equation*}
The method is called temperature scaling and $T$ is a hyper-parameter that will be tuned on the hold-out validation set $D^{dev}_s$
by minimizing the summation of the KL-divergence between the predicted distribution and the true distribution for the examples in the set: $ \sum \kl{\mathbf{y} \| \hat{\mathbf{y}}} $. Note that the method is proposed to be used in confidence calibration~\cite{guo2017calibration}, whereas we use it for calibrating the model outputs to the human label distribution.

\subsection{Distribution Distillation}

Both the Bayesian Inference and Re-Calibration methods do not involve a supervised learning process that is often effective for training models. Here, we consider another method that involves direct training of neural networks called Distribution distillation (reminiscent of model distillation~\cite{hinton2015distilling}).
Distribution distillation consists of three steps. Firstly, we use a ``teacher method'', which can be the Bayesian Inference or Re-Calibration method explained above, to obtain high-quality distribution estimation using $D^{train}$, $D^{dev}$, $D^{dev}_s$. Secondly, we re-label the training set $D^{train}$ with the ``teacher method'' so every example in the training set will be associated with a pseudo soft-label. Finally, we train a new ``student model'' using the relabeled training set. The method is similar to distilling the distribution knowledge of the ``teacher method'' to the final ``student model'' through a large-scale diverse training set.

%% file: 4settings.tex
\section{Experimental Setup}
\label{sec:experimental}

\begin{table}[t]
    \centering
    \small
    \scalebox{0.85}{
    \begin{tabular}{lllll}
    \toprule
    \textbf{Experiments} & $D^{train}$ & $D^{dev}$ & $D^{dev}_s$ & $D^{test}_s$ \\
    \midrule
    \textbf{\chaosNLIa} & 169,654 & 3,059 & 100 & 1,432\\
    \textbf{\chaosNLIs} & 942,854 & 10,000 & 100 & 1,414\\
    \textbf{\chaosNLIm} & 942,854 & 10,000 & 100 & 1,499\\
    \textbf{UNLI} & - & - & - & 2,998 \\
    \textbf{PK2019} & - & - & - & 297 \\ 
    \bottomrule
    \end{tabular}
    }
    \vspace{-5pt}
    \caption{The size of each data split in this work.}
    \vspace{-9pt}
    \label{tab:statistics}
\end{table}

\begin{table*}[t]
    \centering
    \small
    \begin{tabular}{lccccccccc}
    \toprule
    \multirow{2}{*}{\textbf{Model}} 
    & \multicolumn{3}{c}{\textbf{\chaosNLIa}}
    & \multicolumn{3}{c}{\textbf{\chaosNLIs}} 
    & \multicolumn{3}{c}{\textbf{\chaosNLIm}}\\
    \cmidrule(rl){2-4}\cmidrule(rl){5-7}\cmidrule(rl){8-10}
    & \textbf{JSD$\downarrow$} & \textbf{KL$\downarrow$} & \textbf{Acc.$\uparrow$ } & \textbf{JSD$\downarrow$} & \textbf{KL$\downarrow$} & \textbf{Acc.$\uparrow$ } & \textbf{JSD$\downarrow$} & \textbf{KL$\downarrow$} & \textbf{Acc.$\uparrow$ }\\
    \midrule
    \textbf{Chance} & 0.3205 & 0.406 & 0.5052 & 0.383 & 0.5457 & 0.5370 & 0.3023 & 0.3559 & 0.4634\\
    \midrule
    \textbf{Baseline (Mean)} & 0.2033 & 0.8142 & 0.8345           & 0.2160 & 0.4661 & \underline{0.7863}       & 0.3020 & 0.8017 & \underline{0.6324}\\
    \textbf{Baseline (Best)} & 0.2017 & 0.7757 & 0.8317           & 0.2107 & 0.4276 & 0.7822       & 0.2963 & 0.7558 & 0.6318\\
    \textbf{MC Dropout} & 0.1882 & 0.5045 & 0.8251                & 0.1954 & 0.3294 & 0.7845       & 0.2725 & 0.5812 & 0.6231\\
    \textbf{Deep Ensemble} & 0.1941 & 0.6574 & \underline{0.8359}             & 0.2087 & 0.4212 & \textbf{0.7942}       & 0.2926 & 0.7319 & 0.6264 \\
    \midrule
    \textbf{Re-calibration (Oracle)} & 0.1663 & 0.1613 & 0.8345    & 0.1866 & 0.1730 & 0.7863       & 0.2007 & 0.1872 & 0.6324\\
    \textbf{Re-calibration} & \underline{0.1663} & \textbf{0.1615} & 0.8345             & \underline{0.1889} & \textbf{0.1733} & \underline{0.7863}       & \underline{0.2015} & \textbf{0.1873} & \underline{0.6324}\\
    \textbf{MC Dropout (Opt. Rate)} & 0.2046 & 0.3049 & 0.7629                & 0.1970 & 0.2145 & 0.7474       & 0.2525 & 0.3296 & 0.4981 \\
    \textbf{Dist. Distillation} & \textbf{0.1591} & \underline{0.1647} & \textbf{0.8365}                   & \textbf{0.1812} & \underline{0.1802} & 0.7840       & \textbf{0.1969} & \underline{0.1881} & \textbf{0.6374}\\
    \midrule
    \textbf{Human~\cite{nie2020can}} & 0.0421 & 0.0373 & 0.97 & 0.0614 & 0.0411 & 0.94 & 0.0695 & 0.0381 & 0.86\\
    \bottomrule
    \end{tabular}
\vspace{-3pt}
    \caption{Distribution estimation performances on ChaosNLI. $\downarrow$ indicates smaller value is better. $\uparrow$ indicates larger value is better. For each column, the best values are in bold and the second best values are underlined.}
\vspace{-3pt}
    \label{tab:main}
\end{table*}

\subsection{Datasets}
\label{sec:datasets}
We consider the following two NLI-related tasks in our experiments: NLI, and abductive commonsense reasoning.
As described in Sec.~\ref{sec:task}, we need to make the split for $D^{train}$, $D^{dev}$, $D^{dev}_s$, and $D^{test}_s$ for each task.
We use ChaosNLI~\cite{nie2020can} as the data source for $D^{dev}_s$ and $D^{test}_s$ since every example in ChaosNLI are associated with high quality 100 human-annotated labels.\footnote{As explained in Sec.~\ref{sec:motivation}, ChaosNLI is collected with rigid quality control and manual examination reveals that most annotation disagreement results in the actual opinion discrepancy between annotators rather than errors.} Following~\cite{nie2020can}, for each task, we calculate the soft-label for $D^{dev}_s$ and $D^{test}_s$ by using the 100 labels for each example in ChaosNLI. We sampled 100 examples from ChaosNLI and use them for $D^{dev}_s$ and all the other example in ChaosNLI are used for $D^{test}_s$.
We use the training set of SNLI~\cite{bowman2015large} and MNLI~\cite{williams2018broad} as $D^{train}$ for the NLI task and the training set of \abdnli~\cite{DBLP:conf/iclr/BhagavatulaBMSH20} as $D^{train}$ for the abductive reasoning task. 
We use SNLI-test, MNLI-dev-mismatch, and \abdnli-test as the $D^{dev}$.\footnote{The examples in ChaosNLI used in $D^{test}_s$ are mostly from the development splits of the original dataset. Therefore, we need to modify the original dev and test split in this work.}
Additionally, we use the dataset (\pavnli) collected in~\newcite{pavlick2019inherent} as a generalization test set since it contains NLI examples from a different set of domains from MNLI and SNLI.\footnote{There is no official name for the data in~\cite{pavlick2019inherent}. For simplicity, we name it \pavnli.} Note that each example in PK2019 is labeled by 100 annotators with graded labeling schema and we converted the graded labels to 3-way labels (the same format as ChaosNLI) following the conversion guidelines in \newcite{pavlick2019inherent}.
The sizes of each split are in Table~\ref{tab:statistics}, and another Table summarizing the split details here is in the Appendix~\ref{app:dataset}.
Moreover, to understand how well the distribution estimation method can capture individual graded plausibility judgements, we also test our methods on UNLI~\cite{chen-etal-2020-uncertain} where each example is annotated with one single graded label denoting a continuous plausibility score.
For both UNLI and PK2019, we again removed the examples that appeared in our training or development set. The resulting PK2019 dataset used in this work only contains examples from RTE2~\cite{dagan2005pascal}, DNC~\cite{poliak2018collecting} and JOCI~\cite{zhang2017ordinal}.

\subsection{Metrics}
We report the accuracy on the majority label and the KL-divergence and JS-distance (JSD) between the predicted distribution and the soft distribution. On UNLI, we report the Pearson correlation $r$ and the Spearman correlation $\rho$ between the provided graded label and the predicted entailment probability, following the original UNLI setup.\footnote{We do not report the MSE metric for UNLI since their label represents slightly different meanings as our output, our model is not expected to predict the same value as the target.} On PK2019, we report the same metrics as ChaosNLI.

\subsection{Implementation Details}
All the models in this work are built on RoBERTa-Large~\cite{liu2019roberta}. We use the accuracy on the development set ($D^{dev}$) for model selection. We run each model with 10 seeds and report the mean. Additionally, for the baseline experiments, we also report the best performance over 10 different runs. All models are trained using the default dropout rate (0.1) for RoBERTa-Large models.
Hyperparameter details are in the Appendix~\ref{app:hyperparameter}.

%% file: 5results.tex
\begin{table}[t]
    \centering
    \small
    \scalebox{0.75}{
    \begin{tabular}{lccccc}
    \toprule
    \multirow{2}{*}{\textbf{Model}} 
    & \multicolumn{2}{c}{\textbf{UNLI}}
    & \multicolumn{3}{c}{\textbf{PK2019}} \\
    \cmidrule(rl){2-3}\cmidrule(rl){4-6}
    & \textbf{\textit{r}$\uparrow$} & \textbf{{$\rho$}$\uparrow$}  & \textbf{JSD$\downarrow$} & \textbf{KL$\downarrow$} & \textbf{Acc.$\uparrow$ } \\
    \midrule
    \textbf{Baseline (Mean)} & 0.5486 & 0.6421  & 0.2858 & 0.6725 & \textbf{0.6445} \\
    \textbf{MC Dropout}      & 0.5585 & 0.6281  & 0.2699 & 0.5089 & 0.6273 \\
    \textbf{Re-Calibration (S)}   & 0.6344 & 0.6288  & \textbf{0.2469} & \textbf{0.2926} & \textbf{0.6445} \\
    \textbf{Re-Calibration (M)} & \textbf{0.6577} & \textbf{0.6641} & 0.2581 & \textbf{0.2926} & \textbf{0.6445} \\
    \midrule
    \textbf{Train on UNLI} & 0.6762 & 0.6806 & - & - & - \\
    \bottomrule
    \end{tabular}
    }
\vspace{-6pt}
    \caption{Generalization performances on UNLI~\cite{chen-etal-2020-uncertain} and PK2019~\cite{pavlick2019inherent}. The bracket on the right of Re-Calibration denotes the data for $D^{dev}_s$. S=SNLI, M=MNLI. }
\vspace{-3pt}
    \label{tab:generalization}
\end{table}

\begin{table}[t]
    \centering
    \small
    \scalebox{0.85}{
    \begin{tabular}{lcccc}
    \toprule
    \multirow{2}{*}{\textbf{Re-Calibration Data}} 
    & \multicolumn{2}{c}{\textbf{\chaosNLIa}}
    & \multicolumn{2}{c}{\textbf{\chaosNLIm}}\\
    \cmidrule(rl){2-3}\cmidrule(rl){4-5}
    & \textbf{JSD$\downarrow$} & \textbf{KL$\downarrow$} & \textbf{JSD$\downarrow$} & \textbf{KL$\downarrow$} \\
    \midrule
    \textbf{$\left|D^{dev}_s\right|=100$} & 0.1663 &
    \textbf{0.1615}     & 0.2015 & \textbf{0.1873} \\
    \midrule
    \textbf{$\left|D^{dev}_s\right|=10$}  & \textbf{0.1570} & 0.1973       & \textbf{0.1962} & 0.1940  \\
    \midrule
    No soft label
    \textbf{} & 0.1738 & 0.1630 & 0.2347 & 0.3704 \\
    \bottomrule
    \end{tabular}
    }
\vspace{-6pt}
    \caption{Re-Calibration results with different $D^{dev}_s$.}
\vspace{-9pt}
    \label{tab:calib}
\end{table}

\begin{figure}[t]
	\centering
    \includegraphics[width=0.45\textwidth]{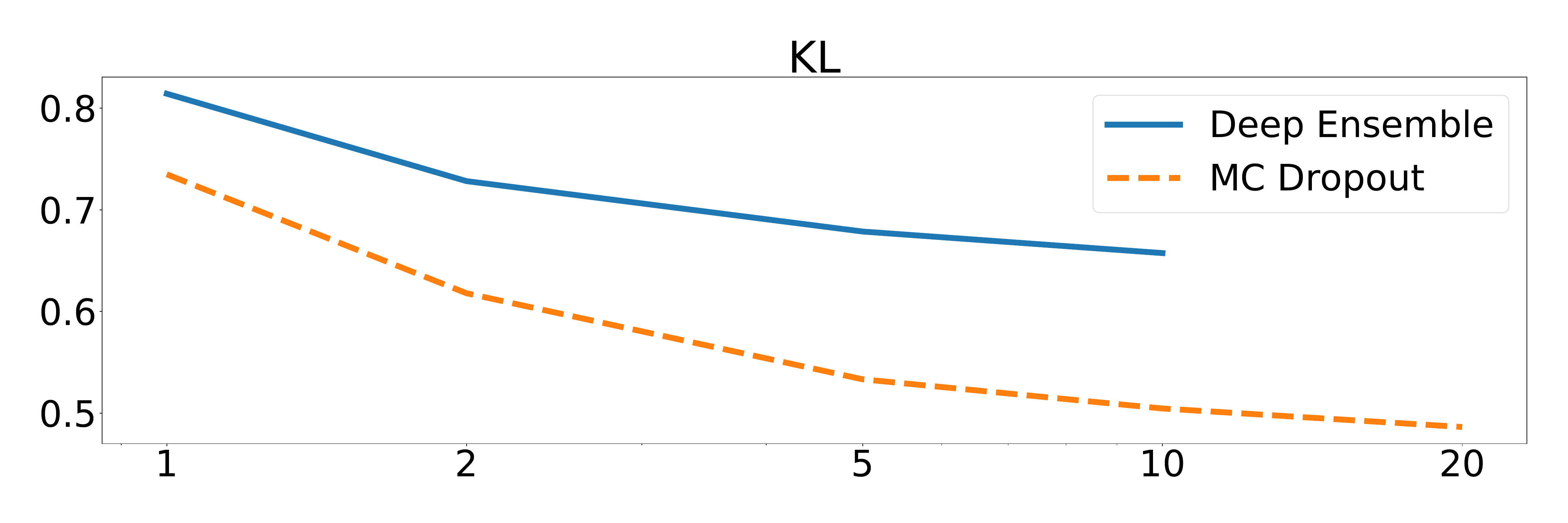}
\vspace{-6pt}
 \caption{KL divergence of MC Dropout and Deep Ensemble with different numbers of model samples.}
\vspace{-3pt}
 \label{fig:npass}
\end{figure}

\section{Results}
\label{sec:results}
The performances of different distribution estimation methods are shown in Table~\ref{tab:main}. The first group (row 2-5) presents the results that do not make use of the soft-labeled data $D^{dev}_s$, while the second group (row 6-8) uses $D^{dev}_s$. Table~\ref{tab:generalization} shows the performance of distribution estimation methods on the out-of-domain test set and the performance of predicting individual graded plausibility judgments. In what follows, we explain the main findings.

\paragraph{MC Dropout is the most preferable method without additional soft-labeled data.}
The first thing we can observe from the first group (row 2-5) in Table~\ref{tab:main} is that both MC Dropout and the Deep Ensemble outperform baselines on all the metrics. 
More importantly, MC Dropout substantially outperforms Deep Ensemble in all KL and JSD columns, with a slight drop on Accuracy. Notice that the MC Dropout results reported in this group are obtained by using the default dropout rate of RoBERTa-Large models (0.1), without tuning on any additional data. The advantage of MC Dropout over Deep Ensemble is different from previous works~\cite{lakshminarayanan2017simple}, and we suspect that this is attributed to the fine-tuning regime of BERT-based models, causing the models in the ensemble to be less diverse. Note that compared to Deep Ensemble, MC dropout
does not require training multiple models.

\paragraph{Further improvement can be obtained by using soft-labeled data, but still below estimated human upper-bound.\footnote{We refer the readers to \citet{nie2020can} for details about the estimation of human upper-bound performance.}}
From the second group (row 6-8) of Table~\ref{tab:main}, we can see further improvement over the Bayesian methods by using additional 100 soft-labeled data $D^{dev}_s$.
For example, on \chaosNLIa, Re-Calibration achieves 0.1615 KL (MC Dropout get 0.5045) and 0.1663 JSD (MC Dropout 0.1882). 
We also include an upper-bound Re-Calibration result by directly applying this method on the test set ($>1000$ examples), but get very close performance to the result with only 100 labels, showing that Re-Calibration is label-efficient. 
With an additional set of soft-labeled data $D^{dev}_s$, we can also tune the optimal dropout rate of MC Dropout.\footnote{In this experiment, the tuning is done by a linear search through 0.0 to 1.0 with step size 0.05. For \chaosNLIa, the searched optimal dropout rate is 0.25, and the value for \chaosNLIs and \chaosNLIm is 0.25 and 0.3 respectively.} The results are shown in the table with name MC Dropout (Opt. Rate). The JSD and KL performance after tuning are substantially higher than the original MC Dropout, however, there is also a substantial decrease on the overall accuracy, and overall this method does not outperform Re-Calibration.
Additionally, Distribution Distillation\footnote{We use Re-Calibration as its teacher method.} only gives slightly better JSD than Re-Calibration, with additional computational cost of retraining the model on the whole relabeled training set, indicating that directly applying Re-Calibration is more efficient. Lastly, the best results here are still below estimated human upper-bound, leaving huge room for improvements.

\paragraph{In-domain improvements hold on the out-of-domain set.}
Table~\ref{tab:generalization} shows the direct generalization results on PK2019 of the models trained on SNLI and MNLI. All the improvements on the in-domain test sets, including MC Dropout over the baseline and the Re-calibration over the MC Dropout, still hold on the out-of-domain examples in the PK2019 test set. Although the out-of-domain scores are generally lower than the in-domain scores in Table~\ref{tab:main}, MC Dropout and Re-Calibration can still bring substantial improvements over the baselines without any PK2019 training data. 

\paragraph{Correlation exists between opinion distribution and graded individual judgement.}
As explained in Sec.~\ref{sec:motivation} and~\ref{sec:datasets}, the distribution of human opinions on NLI examples is different from the individual graded plausibility judgement. In Table~\ref{tab:generalization}, we compare the entailment probability output by the distribution estimation method with the graded plausibility scores in UNLI. Although MC dropout and Re-Calibration method under-perform the baseline on Spearman correlation, Re-Calibration can still greatly improve the Pearson correlation $r$. More importantly, our best distribution estimation method without using any UNLI data is noticeably comparable to the reported numbers in UNLI~\cite{chen-etal-2020-uncertain} using a fine-tuned model on in-domain UNLI data. This hints at a certain correlation between opinion distributions and graded individual judgements, consistent with our intuition regarding the interpretation of the labels.

%% file: 6analysis.tex
\section{Ablation \& Analysis}
\label{sec:analysis}

\subsection{Ablations for Re-Calibration}
In the previous section, we showed the effectiveness of Re-Calibration in predicting human opinion distribution by explicitly utilizing extra soft-labeled data $D^{dev}_s$. To get a better sense of what contributes to the performance, we make two ablations on $D^{dev}_s$: (1) reducing the size of $D^{dev}_s$ from 100 to 10; (2) using only the majority class as hard labels in $D^{dev}_s$ rather than the whole label distribution as soft labels for Re-Calibration. Table~\ref{tab:calib} shows the results.\footnote{See Appendix~\ref{app:recalibration} for full results including \chaosNLIs.}
We can observe that with only 10 examples, Re-Calibration can already achieve good performance, with slightly worse KL but slightly better JSD.\footnote{The diverse trend is because that the re-calibration is conducted only using the KL metric, but the temperature with the best KL metric does not lead to the best JSD.}
However, using only the hard labels gives significantly worse scores than using the soft labels on \chaosNLIm, indicating the necessity of extra annotations in human distribution modeling.

\subsection{Sample Sizes in Bayesian Method}
In both MC Dropout and Deep Ensemble methods, the distribution is approximated by sampling. To understand how the number of samples will influence the results, we present the result for both methods with different numbers of samples ($k$) on \chaosNLIa in Figure~\ref{fig:npass}. We can see that while a larger number of samples will lead to better distribution estimation results on KL, the gain is gradually diminished (even with the log-scale x-axis in Figure~\ref{fig:npass}). Similar trends can also be seen on JSD and Accuracy and the figures are in the Appendix~\ref{app:samplesize}. Considering practical constraints on inference time and computational budget that prohibit a very large number of samples, in our experiments, numbers around 10 is a sweet point between good performance and an acceptable computational budget.

\subsection{Distribution Prediction Example}
Table~\ref{tab:example} shows an example from MNLI with the prediction distribution from both MC Dropout and Re-Calibration. We can see that one-third of humans believe the label should be entailment and two-third as neutral. It is commonsense that clerking for an honorable District Court can be a really rewarding experience, though the premise does not explicitly say so. The MC Dropout method underestimates such a factor and gives more than 80\% for the neutral label. Notably, although Re-Calibration method predicts a smoother distribution that resembles human distribution more, it ends up erroneously increasing the probability for the inexplicable contradiction label. Such observation that MC Dropout tends to overlook the disagreement and Re-Calibration can sometimes produce smooth distribution but with erroneously high probabilities is common and should be taken into consideration before practical deployment. More error-analysis examples and a more detailed comparison of the predictions of distribution prediction methods on the whole-dataset level is in the Appendix~\ref{app:analysis_example}, \ref{app:analysis_distribution}.

\begin{table}[t]
\centering
\scalebox{0.60}{
\begin{tabular}{rp{22em}}
\toprule
\textbf{Premise} & Professor Rogers began her career by clerking for The Honorable Thomas D. Lambros of the United States District Court for the Northern District of Ohio.\\
\textbf{Hypothesis} & Her career benefited from being a clerk to Thomas.\\
\midrule
\multicolumn{2}{c}{\textit{ Prediction (Entailment / Neutral / Contradiction)}} \\
\midrule
\textbf{Human Distribution} & 0.33 / 0.66 / 0.01 \\
\textbf{MC Dropout} & 0.1617 / 0.8362 / 0.0021 \\
\textbf{Re-Calibration} & 0.3063 / 0.5902 / 0.1035 \\
\bottomrule
\end{tabular}
}
\vspace{-6pt}
\caption{An example of prediction distribution and human ground truth in MNLI.}
\vspace{-6pt}
\label{tab:example}
\end{table}

%% file: 7related.tex
\section{Related Work}

Inherent disagreement and ambiguity in NLP annotations has a long history~\cite{poesio2005reliability,zeman2010hard} involving tasks like coreference resolution~\cite{poesio2008justified,poesio2019crowdsourced,li2020neural}, POS-tagging~\cite{zeman2010hard, plank2014learning, plank2016multilingual}, semantic frame disambiguation~\cite{dumitrache2019crowdsourced}, humorousness prediction~\cite{simpson2019predicting}, etc. 
Most previous works design methods to predict one single gold label by aggregating the noisy information~\cite{dawid1979maximum, hovy2013learning, rodrigues2017learning, paun2018probabilistic, braylan2020modeling, Uma2021beyond}. On the contrary, following the recent definition in NLI works \cite{chen-etal-2020-uncertain, pavlick2019inherent, nie2020can}, we directly try to predict  distribution labels that accurately reflect the opinion of a large population. \citet{peterson2019human} is most similar to us in label definition, and studied the advantage of using distribution labels in image classification.

Uncertainty and calibration have also been studied in various NLP models, from traditional structured prediction models~\cite{nguyen2015posterior}, to seq-to-seq models~\cite{ott2018analyzing, kumar2019calibration,xu2020understanding} and transformers~\cite{desai2020calibration}. 
\citet{gantt-etal-2020-natural} suggests a constructive view of NLI modeling in which the prediction is explicitly grounded on annotator identifiers, incorporating the annotator random effects. \citet{zhang2021identifying} propose an ensemble-based framework that can identify examples with high label disagreement. \citet{xiao2019quantifying} shows that explicitly modeling the uncertainty can improve performance, and \citet{wang2020inference} propose a label smoothing method that improves calibration for NMT.
Instead of aiming for a better uncertainty, our work uses multiple uncertainty estimation methods for more accurate distribution prediction.
Concurrently, \citet{meissner-etal-2021-embracing} explores training models directly on the multiple labels from each annotator in SNLI and MNLI, \citet{zhang2021capturing,zhang2021learning} also leverage distribution labels in the model development process and explore training methods combining the supervision signal of one-hot and distributional labels.
In comparison, our work studies additional Bayesian estimation methods and provides a detailed discussion on why and how modeling distribution labels is beneficial for NLU, including the motivation, nuances, and evaluation standardization.

%% file: 8conclusion.tex
\section{Conclusion}
We introduce distributed NLI -- an extension of NLI with a new goal of predicting human opinion distribution. We show that several distribution estimation methods can capture such distributions more effectively than softmax, but the best results are still far below the estimated upper-bound. We analyze the properties and weaknesses of the methods, highlight the importance of the task, and encourage future work on developing better models for estimating the human opinion distribution.

%% file: appendix.tex
\appendix
\newpage

\begin{table*}[t]
    \centering
    \small
    \scalebox{0.98}{
    \begin{tabular}{lllll}
    \toprule
    \textbf{Experiments} & \textbf{Train} $D^{train}$ & \textbf{Dev} $D^{dev}$ & \textbf{Soft Dev} $D^{dev}_s$ & \textbf{Test} $D^{test}_s$ \\
    \midrule
    \textbf{\chaosNLIa} & $\alpha$NLI\textsuperscript{train} (169654) & $\alpha$NLI\textsuperscript{Test} (3059) & $\alpha$NLI\textsuperscript{dev} (100) & \chaosNLIa - $D^{dev}_s$ (1432)\\
    \textbf{\chaosNLIs} & SNLI\textsuperscript{train} + MNLI\textsuperscript{train} (942854)& SNLI\textsuperscript{Test} (10000) & SNLI\textsuperscript{dev} (100) & \chaosNLIs - $D^{dev}_s$ (1414)\\
    \textbf{\chaosNLIm} & SNLI\textsuperscript{train} + MNLI\textsuperscript{train} (942854) & $\text{MNLI}^{\text{dev}}_{\text{mismatch}}$ (10000) & $\text{MNLI}^{\text{dev}}_{\text{match}}$ (100) & \chaosNLIm - $D^{dev}_s$ (1499)\\
    \textbf{UNLI} & - & - & - & UNLI (2998) \\
    \textbf{PK2019} & - & - & - & PK2019 (297) \\ 
    \bottomrule
    \end{tabular}
    }
    \caption{Data sources for each split. The corresponding size of each split is shown in the bracket after the source.}
    \label{tab:statistics_app}
\end{table*}

\begin{table*}[t]
    \centering
    \small
    \begin{tabular}{lcccccc}
    \toprule
    \multirow{2}{*}{\textbf{}} 
    & \multicolumn{2}{c}{\textbf{\chaosNLIa}}
    & \multicolumn{2}{c}{\textbf{\chaosNLIs}} 
    & \multicolumn{2}{c}{\textbf{\chaosNLIm}}\\
    \cmidrule(rl){2-3}\cmidrule(rl){4-5}\cmidrule(rl){6-7}
    & \textbf{JSD$\downarrow$} & \textbf{KL$\downarrow$} &  \textbf{JSD$\downarrow$} & \textbf{KL$\downarrow$}  & \textbf{JSD$\downarrow$} & \textbf{KL$\downarrow$} \\
    \textbf{$D^{dev}_s=100$} & 0.1663 & \textbf{0.1615}         & 0.1889 & \textbf{0.1733}      & 0.2015 & \textbf{0.1873} \\
    \textbf{$D^{dev}_s=10$}  & \textbf{0.1570} & 0.1973         & \textbf{0.1744} & 0.1977      & \textbf{0.1962} & 0.1940  \\
    No soft label & 0.1738 & 0.1630 &  0.2008 & 0.3667 & 0.2347 & 0.3704 \\
    \bottomrule
    \end{tabular}
    \caption{Re-calibration performances with different types of labels. $\downarrow$ indicates smaller value is better. $\uparrow$ indicates larger value is better. For each column, the best values are in bold.}
    \label{tab:calib_app}
\end{table*}

\section{Dataset Split Details}
\label{app:dataset}
The details of dataset split, including the source of the data and the corresponding size in Table~\ref{tab:statistics_app}. The UNLI data can be downloaded at \url{https://nlp.jhu.edu/unli/}. The \pavnli data is at \url{https://github.com/epavlick/NLI-variation-data}. The ChaosNLI data is at \url{https://github.com/easonnie/ChaosNLI}.

\section{Hyperparameter Details}
\label{app:hyperparameter}
All the models in this work are built on RoBERTa-Large~\cite{liu2019roberta}. For both NLI and \abdnli tasks, we fine-tune our model with peak learning rate 5e-6, warm-up ratio 0.1 and linear learning rate decay. We use a batch size of 32. We train the NLI model for 2 epochs, and the \abdnli model for 3 epochs. We always use the accuracy on the development set ($D^{dev}$) for model selection. All our experiments are conducted on a single server with 4 GTX 1080Ti GPUs.

\section{Full Re-Calibration Ablation}
\label{app:recalibration}
The full results of Re-Calibration ablations is shown in Table~\ref{tab:calib_app}. We can see on all three subsets of ChaosNLI, Re-Calibration always achieves good performance even with as few as 10 additional distribution labels; and using 100 distribution labels always significantly outperforms using 100 hard labels without any distribution information.  

\begin{figure*}[t]
	\centering
    \includegraphics[width=1.00\textwidth]{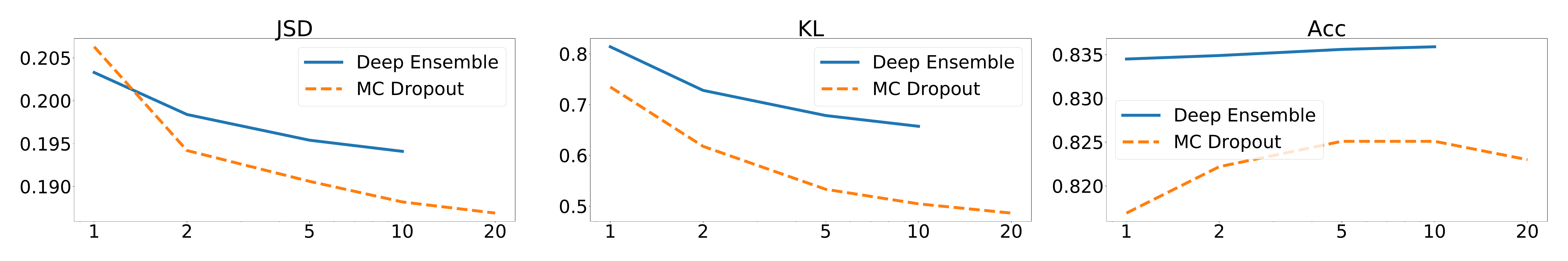}
 \caption{Performances of MC Dropout and Deep Ensemble with different numbers of model samples.}
 \label{fig:npass_app}
\end{figure*}

\section{The Effect of Sample Size}
\label{app:samplesize}
Figure~\ref{fig:npass_app} shows model performances on all three metrics (JSD, KL and Accuracy) with different sample sizes. We can observe similar trends on the KL metric as discussed in the main paper. While a larger number of samples usually leads to better performance, the gain is gradually diminished.

\begin{figure*}[t]
	\centering
    \includegraphics[width=1.00\textwidth]{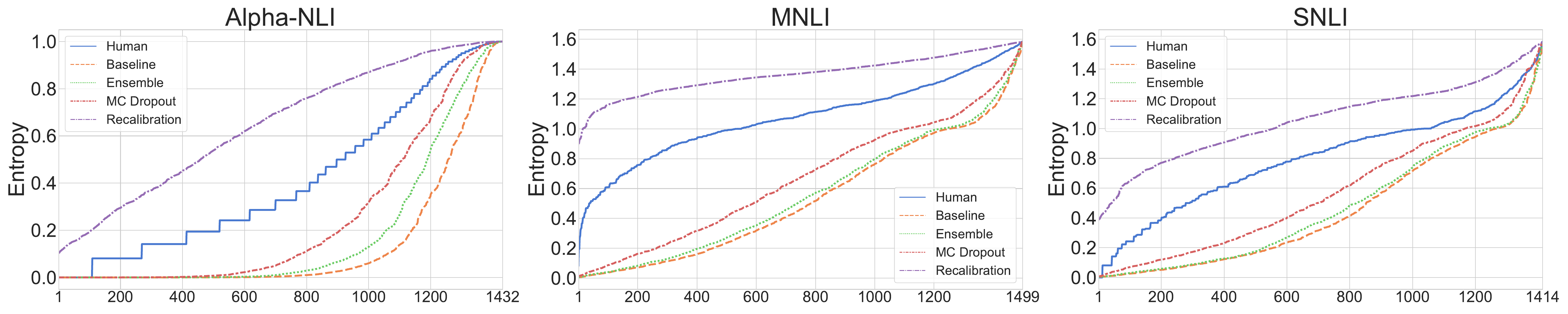}
 \caption{Entropy quantile curve on ChaosNLI. Each point in this figure represents model's prediction on one example. y-axis is the entropy value, and x-axis is the prediction's index in a sequence of examples sorted in the increasing-entropy order.}
 \label{fig:entropy_app}
\end{figure*}

\begin{table}[t]
\tiny
\centering
	\begin{tabular}{p{21em}p{9em}}
	\toprule
	\textbf{Premise} & \textbf{Hypothesis} \\
	\midrule
Case Description: Some dark night a policeman walks down a street, apparently deserted; but suddenly he hears a burglar alarm, looks across the street, and sees a jewelry store with a broken window. Then a gentleman wearing a mask comes crawling out through the broken window, carrying a bag which turns out to be full of expensive jewelry.\footnotemark{} & The gentleman is dishonest and guilty for stealing.\\
    \midrule

The \textit{Off Fossil Fuels for a Better Future Act} lays out that by 2035: (1) 100\% of electricity must be generated from clean energy resources, (2) 100\% of vehicle sales from manufacturers must be zero-emission vehicles, and (3) 100\% of train rail lines and train engines must be electrified. & Passing the bill means embracing clean energy sources for the good of sustainable development.\\
	\bottomrule
	\end{tabular}
	\caption{\label{tab:decision_making} Examples where AI-aided human decision making can be formulated as an NLI problem.} 
\end{table}

\section{Additional Motivation and Positioning of Distributed NLI}
\label{app:motivation}
\subsection{Potential Applications of Distributed NLI}
\label{app:application}

In order for NLU models to aid humans in decision-making, it is important for NLI models to output a valid distribution and to capture the opinions of the minority sub-populations. We include two example inputs in such situations in Table~\ref{tab:decision_making}.

\begin{table*}[t]
\centering
\scalebox{0.70}{
\begin{tabular}{rp{44em}}
\toprule
\textbf{Premise} & you want to punch the button and go \\
\textbf{Hypothesis} & You don't want to push the button lightly, but rather punch it hard. \\
\midrule
\multicolumn{2}{c}{\textit{ Prediction (Entailment / Neutral / Contradiction)}} \\
\midrule
\textbf{Original Annotation} & 0.48 / 0.45 / 0.07 \\
\textbf{Incorrect Labels} & Contradiction \\
\textbf{Reasons for Incorrect Labels} & There is no sufficient evidence in the premise indicating "you also want to push the button lightly". \\
\midrule \midrule
\textbf{Premise} & The tomb guardian will unlock the gate to the tunnel and give you a candle to explore the small circular catacomb, but for what little you can see, it is hardly worth the effort. \\
\textbf{Hypothesis} & The tomb garden can give you a thorough tour of the catacombs. \\
\midrule
\multicolumn{2}{c}{\textit{ Prediction (Entailment / Neutral / Contradiction)}} \\
\midrule
\textbf{Original Annotation} & 0.10 / 0.14 / 0.76 \\
\textbf{Incorrect Labels} & Entailment \\
\textbf{Reasons for Incorrect Labels} & The premise mentions "tomb guardian" instead of "tomb garden", so it should not be entailment. \\
\bottomrule
\end{tabular}
}
\caption{Examples of wrong annotations on the $D^{dev}_s$ split of \chaosNLIm. }
\label{tab:ex_err}
\end{table*}

\begin{table*}[t]
\centering
\scalebox{0.70}{
\begin{tabular}{rp{44em}}
\toprule
\textbf{Premise} & They said that (1) agencies need to be able to design their procedures to fit their particular circumstances (e.g.\\
\textbf{Hypothesis} & The authors of the recently introduced bill stated each agency would be required to match their operational methods to their particular situations.\\
\midrule
\multicolumn{2}{c}{\textit{ Prediction (Entailment / Neutral / Contradiction)}} \\
\midrule
\textbf{Human Distribution} & 0.58 / 0.30 / 0.12 \\
\textbf{Baseline} & 0.002 / 0.997 / 0.001 \\
\textbf{Reasons for Ambiguity} & Based on different understanding of the phrases "need to be able to" in the premise, this sentence pair can have different labels. \\
\midrule \midrule
\textbf{Premise} & What changed? \\
\textbf{Hypothesis} & Nothing changed. \\
\midrule
\multicolumn{2}{c}{\textit{ Prediction (Entailment / Neutral / Contradiction)}} \\
\midrule
\textbf{Human Distribution} & 0.04 / 0.76 / 0.20 \\
\textbf{Baseline} & 0.001 / 0.007 / 0.992 \\
\textbf{Reasons for Ambiguity} & In different contexts, the question in the premise can imply different meanings. \\
\bottomrule
\end{tabular}
}
\caption{Examples of prediction distribution of the baseline model and human ground truth in MNLI.}
\label{tab:ex_app1}
\end{table*}

\begin{table}[t]
    \centering
    \small
    \scalebox{0.80}{
    \begin{tabular}{lcccc}
    \toprule
    \multirow{2}{*}{\textbf{Model}} 
    & \multicolumn{2}{c}{\textbf{JSD$\downarrow$}}
    & \multicolumn{2}{c}{\textbf{KL$\downarrow$}} \\
    \cmidrule(rl){2-3}\cmidrule(rl){4-5}
    & \textbf{Original} & \textbf{Corrected}  & \textbf{Original} & \textbf{Corrected} \\
    \midrule
    \textbf{Baseline (Mean)} & 0.3053 & 0.3039  & 0.8383 & 0.8343  \\
    \textbf{MC Dropout}      & 0.2649 & 0.2653  & 0.5851 & 0.5839 \\
    \textbf{Deep Ensemble}      & 0.2956 & 0.2941  & 0.7775 & 0.7709 \\
    \textbf{Re-Calibration} & \textbf{0.1983} & \textbf{0.2079} & \textbf{0.1859} & \textbf{0.1983}  \\
    
    \midrule
    \bottomrule
    \end{tabular}
    }
    \caption{Performances difference on the $D^{dev}_s$ split of \chaosNLIm. }
    \label{tab:performance_diff}
\end{table}

\subsection{Analysis of Annotation Quality}
\label{app:quality}

We manually examined the label correctness of the 100 examples in the $D^{dev}_s$ split of \chaosNLIm. Due to the careful quality control over label collection, only a very limited set of the labels are incorrect. Out of the 100 examples in the examined subset, only 4 examples contain more than 10 error annotations and no example contain more than 16 error annotations.
In Table~\ref{tab:ex_err}, we show two examples of incorrect label annotations in the $D^{dev}_s$ split of \chaosNLIm. While both examples do contain a certain level of semantic ambiguities, with careful reasoning, we do not find sufficient evidence to make the "contradiction" or "entailment" judgement in those cases respectively, hence we view these labels as error annotations.

 We also verified these incorrect labels will not substantially impact the results in this paper. For these 100 examples, we removed all the incorrect labels with more than 5 annotations and created a \textbf{corrected} label set. The performance difference between the \textbf{original} annotations and the \textbf{corrected} annotations can be seen in Table~\ref{tab:performance_diff}. We can see only marginal performance difference is shown for the Baseline, MC Dropout and Deep Ensemble variants. The performance difference for the Re-Calibration variant is slighter larger due to the fact that these labels are also directly used in the temperature calibration process, but it also only leads to a relatively small difference around 0.01. Furthermore, using either the original or the corrected labels, the order of more effective methods (Re-Calibration > MC Dropout > Deep Ensemble > Baseline) always holds.

\subsection{Predicting Label Distribution from Deterministic Labels}
\label{app:deterministic}
Another question around the feasibility of the Distributed NLI is whether model can learn label distributions if only deterministic labels are possible. Here we prove it is definitely possible if the deterministic labels are annotated by individual annotators.\footnote{Annotations from datasets like MNLI and SNLI can still be roughly viewed as labels from individual annotators with an additional voting-based filtering methods that filters out noisy labels using voting among 5 annotators.} Specifically, if we assume all the training inputs $x$ are randomly sampled from a dataset $D$, and each corresponding label $y$ is provided from a random annotator $a$ from a set of annotators $A$. Then, on the training set, the model is trained to minimize $\displaystyle \mathop{\mathbb{E}}_{x\in D}\mathop{\mathbb{E}}_{a\in A}\kl{y_a(x) \| P(x)}$. Specifically, assuming the model parameter is $\theta$, the optimization target is: 
\begin{align*}
    &\argmin_\theta \displaystyle \mathop{\mathbb{E}}_{x\in D}\mathop{\mathbb{E}}_{a\in A}\kl{y_a(x) \| P_\theta(x)} \\
    =&\argmin_\theta \displaystyle \mathop{\mathbb{E}}_{x\in D}\mathop{\mathbb{E}}_{a\in A}\sum_{j} y_a(x)_j\frac{\log y_a(x)_j}{\log P_\theta(x)_j} \\
    =&\argmin_\theta \displaystyle \mathop{\mathbb{E}}_{x\in D}\mathop{\mathbb{E}}_{a\in A}\sum_{j} -y_a(x)_j \log P_\theta(x)_j \\
    =&\argmin_\theta \displaystyle \mathop{\mathbb{E}}_{x\in D} \sum_{j} -(\mathop{\mathbb{E}}_{a\in A} y_a(x))_j \log P_\theta(x)_j \\
    =&\argmin_\theta \displaystyle \mathop{\mathbb{E}}_{x\in D} \sum_{j} (\mathop{\mathbb{E}}_{a\in A} y_a(x))_j\frac{\log (\mathop{\mathbb{E}}_{a\in A} y_a(x))_j}{\log P_\theta(x)_j} \\
    =&\argmin_\theta \displaystyle \mathop{\mathbb{E}}_{x\in D} \kl{\mathop{\mathbb{E}}_{a\in A} y_a(x) \| P_\theta(x)}
\end{align*}
, where $j$ is each dimension of the output label. 
Hence, even with deterministic labels, the model still achieves the best performance if and only if when given an example $x$, $P_\theta(x) =\displaystyle \mathop{\mathbb{E}}_{a\in A }y_a(x)$, where the model correctly predicts the distribution of all labels.

\section{More Analysis on Distribution Prediction Examples}
\label{app:analysis_example}
\addtocounter{footnote}{-1}
\stepcounter{footnote}\footnotetext{The example is from \newcite{jaynes2003probability}.}
In this section, we provided more prediction examples and a more comprehensive analysis in addition to the examples shown in the Ablation \& Analysis section in the main paper. Specifically, we focus on analyzing what are the worst-prediction examples produced by current models.

For each model variant, we checked the performance on the $D^{dev}_s$ split on \chaosNLIm and focused on the examples with the largest KL-divergence (worst-prediction examples). For the baseline, we again noticed the trend that models being over-confident on examples with substantial ambiguity. We show two examples in Table~\ref{tab:ex_app1}. In both of these cases, the model fails to capture the label distribution caused by subtle phrase relationships or under-specified meaning depending on the context, etc.
As shown in the results section in the main paper, Such over-confidence can be partially alleviated by the Bayesian uncertainty estimation methods (e.g., MC-Dropout and Deep Ensemble) and by the Re-calibration methods. However, most of the top 10 worst-prediction examples of the baseline variant still remain in the top 10 worst-prediction list of the Bayesian and Re-calibration approaches. This observation is possibly due to the limited improvement of Bayesian approaches and the incapability of Re-calibration methods to correct totally wrong predictions, hence showing current models' inherent incapability to capture these distributions. We also encourage future work to explore the connection between the model's incapability to capture distribution labels and the model's tendency to focus on artifact features.

\section{Prediction Difference in Bayesian Inference and Re-Calibration}
\label{app:analysis_distribution}
We show that both Bayesian Inference and Re-Calibration can achieve better JSD and KL scores in the main paper. In order to investigate the difference between the predictions produced by the two methods, we conduct the following analysis.
Firstly, for each example in the test set, we calculate the entropy for the models outputs as $\label{equ:entropy}
\mathbf{H}\left( \mathbf{p} \right) = - \sum_{i \in \{e, n, c\} }{p_i \log(p_i)} 
$ where $p_i$ is the probability for entailment, neutral, or contradiction. We also calculate the entropy for human using the annotations in ChaosNLI. 
We then sort the entropy and plot their entropy values for each model. The plot is shown in Fig.~\ref{fig:entropy_app}.\footnote{The design of the figure is similar to the Q–Q (quantile-quantile) plot~\cite{gnanadesikan1968probability}, a visualization method to compare two probability distributions by plotting their quantiles against each other. We modify the plot to give an intuitive comparison for all the distribution estimation methods in our study.} 
We can see a large gap between the blue line representing human distribution and the orange dashed line representing the baseline, consistent with previous quantitative findings. While Bayesian inference methods can slightly reduce this gap, there is still large room for improvements. Moreover, the distribution predicted by the Re-Calibration method is noticeably different from the ones given by the MC Dropout, Ensemble, and the baseline method, while the latter three are very similar to each other. Finally, it is worth noting that the line for the Re-Calibration method is above the human line while the other three methods are below the human line. 
This suggests that Re-Calibration method tends to over-predict the disagreement among humans whereas the Bayesian method and the baseline fail to capture some inherent disagreements.